\documentclass[10pt,conference]{IEEEtran}
\IEEEoverridecommandlockouts
\usepackage{cite}
\usepackage{amsmath,amssymb,amsfonts}
\usepackage{algorithmic}
\usepackage{graphicx}
\usepackage{textcomp}
\usepackage{xcolor}
\usepackage{enumitem}
\usepackage{listings}
\usepackage{booktabs}
\usepackage{makecell}
\usepackage{multirow}
\usepackage{tabularx}
\usepackage{soul, color}
\usepackage{balance}
\usepackage[table]{xcolor}
\definecolor{headerblue}{RGB}{214, 235, 250}
\definecolor{rlorange}{RGB}{255, 231, 204}
\def\BibTeX{{\rm B\kern-.05em{\sc i\kern-.025em b}\kern-.08em
    T\kern-.1667em\lower.7ex\hbox{E}\kern-.125emX}}

\newcolumntype{Y}[1]{%
    >{\hsize=#1\hsize\centering\arraybackslash}X%
}

\newcommand{\swecritix}{\textsc{SWE-Critix}}
\newcommand{\swecritixws}{\textsc{SWE-Critix }}

\begin{document}

\title{\swecritix: Post-Training Generalist Verifiers for Assessing Coding-Agent Trajectories}

\author{\IEEEauthorblockN{Jiajun Hu}
\IEEEauthorblockA{
\textit{Leibniz Research Center} \\
\textit{Huawei}\\
Hong Kong SAR, China \\
hu.jiajun@huawei.com}
\and
\IEEEauthorblockN{Chun Yong Chong}
\IEEEauthorblockA{\textit{School of Information Technology} \\
\textit{Monash University Malaysia}\\
Subang Jaya, Malaysia \\
chong.chunyong@monash.edu}
}

\maketitle

\begin{abstract}
Coding agents are increasingly deployed for end-to-end software engineering tasks, where a readily available oracle evaluator that defines the success criteria is often absent. This motivates verifications for coding agents that can reliably judge and explain their behavior. However, verification is non-trivial as coding agents are increasingly capable of solving long-horizon software engineering tasks. Their execution trajectories can be extremely long, containing hundreds of iterative interactions with code repositories and external tools. Existing verifiers either rely on general-purpose LLMs without specialized training or train dedicated critics that predominantly focus on producing scalar rewards, highlighting the relatively limited exploration of training-based generalist verification. We introduce \swecritix, a post-training pipeline for training generalist verifiers of coding-agent trajectories. Given a task and its execution trajectory, the learned verifier predicts the completion label and generates a natural-language analysis that explains its judgment. Our pipeline combines supervised fine-tuning (SFT) with reinforcement learning (RL) and trains a series of 30B models on trajectories from CodeForge-Preview. On a held-out test set, \swecritixws improves classification accuracy from 57\% for the base Qwen3-30B-A3B reasoning model to 66\% after SFT and 78\% after RL. We release our models, datasets, and training code to facilitate further research.

\end{abstract}


\section{Introduction}
Recent advances in large language models (LLMs) have enabled a shift from code generation toward autonomous coding agents that can interact with environments to solve complex software engineering tasks. Rather than producing one-shot code snippets, coding agents can inspect repositories, edit files, execute commands, run tests, interpret feedback, and iteratively refine their solutions through interactions with external tools~\cite{deepseek-harness2026,codex,claudecode}. This agentic paradigm has substantially expanded the scope of automated programming from isolated function synthesis to repository-level issue resolution~\cite{jimenez2024swebench} and long-horizon software development tasks~\cite{deng2025swebenchproaiagents,peng2026lolbenchevaluatingcodingagents}.

As coding agents become more capable and autonomous, reliably verifying their execution trajectories (i.e., step-by-step record of an agent solving a task) becomes increasingly important. In realistic end-to-end tasks, the criteria for successful completion are often not explicitly specified by a readily available ground-truth oracle. A verifier is therefore needed to determine whether an agent has accomplished the task and, when it fails, identify the causes of failure. However, such verification is challenging because coding-agent trajectories can contain hundreds of interleaved user inputs, environment feedback, and model responses, making them substantially more difficult to inspect than a one-shot model response. For instance, trajectories for tasks in LoLBench~\cite{peng2026lolbenchevaluatingcodingagents} span 53--532 turns and contain millions of tokens on average.

To address the challenge, existing work have proposed different techniques from two orthogonal dimensions: verification \emph{target} and \emph{method}, as summarized in Table~\ref{tab:verifier_taxonomy}. The former distinguishes whether a verifier evaluates only the agent's final outcome (e.g., the submitted patch) or also examines the process to detect undesirable intermediate behaviors. The latter distinguishes whether a verifier is constructed with or without training. Training-free methods directly employ general-purpose LLMs or conventional rule-based judges, whereas training-based methods learn dedicated critics, which can be further characterized by whether they produce scalar rewards or richer natural-language assessments.

\begin{table}[t]
    \centering
    \caption{A 2D taxonomy of coding-agent verifiers.}
    \label{tab:verifier_taxonomy}
    \setlength{\tabcolsep}{3.5pt}
    \renewcommand{\arraystretch}{1.12}

    \begin{tabularx}{\columnwidth}{
        @{}
        >{\raggedright\arraybackslash}p{0.20\columnwidth}
        >{\raggedright\arraybackslash}X
        >{\raggedright\arraybackslash}X
        @{}
    }
        \toprule
        & \textbf{Outcome} & \textbf{Process} \\
        \midrule

        \textbf{Training-free}
        &
        Agentless~\cite{xia2024agentlessdemystifyingllmbasedsoftware};
        LLM-as-a-Verifier~\cite{kwok2026llmasaverifiergeneralpurposeverificationframework};
        Trae~\cite{traeresearchteam2025traeagentllmbasedagent};
        Agentic Rubrics~\cite{raghavendra2026agenticrubricscontextualverifiers}
        &
        SWE-PRM\cite{gandhi2025agentsastraycoursecorrectingswe};
        Wink\cite{nanda2026winkrecoveringmisbehaviorscoding};
        TrajDebug\cite{qi2026trajdebugtracingerrorlifecycle};
        SWE-Prime\cite{zheng2026sweprimefewertrajectoriesbetter};
        ProcCtrlBench\cite{he2026procctrlbenchevaluatingprocessleveldefects}
        RETRACE\cite{li2026independentpatchverificationcoding};
        LivePlan\cite{liu2026onlinemonitoringcorrectivesteering};
        AgentLens\cite{sahoo2026agentlensrevealingluckypass};
        
        \\

        \midrule

        \makecell[tl]{
            \textbf{Training-based}\\[0pt]
            \hspace{2.0em}\textit{Scalar}
        }
        &
        SWE-Gym\cite{pan2025trainingsoftwareengineeringagents};
        R2E-Gym\cite{jain2025r2egymproceduralenvironmentshybrid};
        DeepSWE\cite{deepswe2025};
        SWE-RM\cite{shum2025swermexecutionfreefeedbacksoftware};
        Dockerless\cite{zeng2026dockerlessenvironmentfreeprogramverifier};
        SWE-TRACE\cite{han2026swetraceoptimizinglonghorizonswe}
        &
        OpenHands Critic\cite{wang2025-sota-on-swe-bench};
        Zainullina et al.\cite{zainullina2025guidedsearchstrategiesnonserializable};
        Critic Rubrics\cite{wang2026rubricsupervisedcriticsparserealworld}
        \\[4pt]

        \midrule

        \makecell[tl]{
            \textbf{Training-based}\\[0pt]
            \hspace{1.3em}\textit{Generalist}
        }
        &
        Agentic Rubrics\cite{raghavendra2026agenticrubricscontextualverifiers};
        ExecCritic\cite{tao2026execcriticlearntesttest};
        SWE-Review\cite{wang2026swereviewclosingloopissue}
        &
        Gandhi et al.~\cite{gandhi2026steerdontsolvetraining};
        Scout~\cite{rahman2026locatinghiddenfailuresmakes};
        \hl{SWE-Critix}
        \\

        \bottomrule
    \end{tabularx}
\end{table}

\begin{figure*}[htbp]
  \centering
  \includegraphics[width=\textwidth]{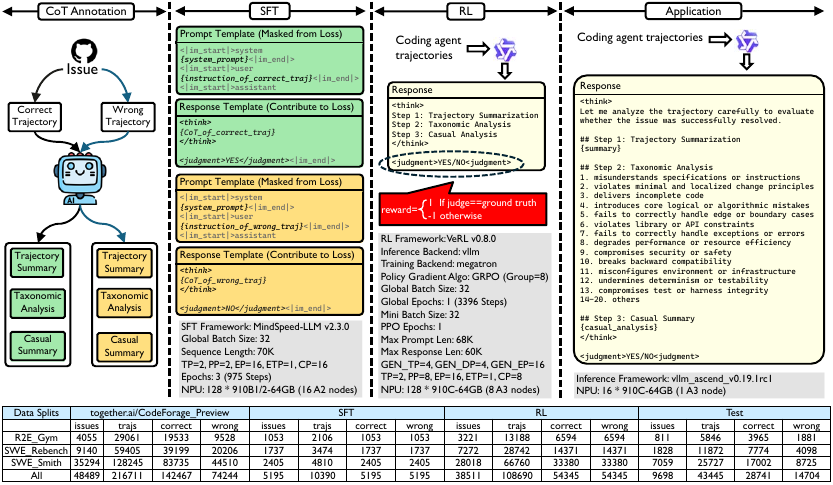}
  \caption{\swecritixws Post-Training Pipeline and Data Splits.}
  \label{fig:overview}
\end{figure*}

This taxonomy reveals that training-based, process-oriented generalist verification remains relatively underexplored. To fill the gap, we propose \swecritixws (Figure~\ref{fig:overview}), a post-training pipeline that combines supervised fine-tuning (SFT) and reinforcement learning (RL) to train a generalist trajectory verifier. Given a task and its execution trajectory, the learned verifier not only determines whether the task has been successfully completed but also justifies the rationale underlying its judgment. Compared with the two existing approaches in this category, \swecritixws differs from them in terms of verification target (planning phase~\cite{gandhi2026steerdontsolvetraining} vs. whole trajectory), annotation strategy (per-step dense labeling~\cite{rahman2026locatinghiddenfailuresmakes} vs. differential analysis), scale (2--8K training samples and 4--8B models~\cite{gandhi2026steerdontsolvetraining,rahman2026locatinghiddenfailuresmakes} vs. 119K and 30B), learning algorithms (DPO~\cite{gandhi2026steerdontsolvetraining} vs. GRPO), and reward design (process reward~\cite{rahman2026locatinghiddenfailuresmakes} vs. outcome reward).

Using this pipeline, we train a Qwen3-30B MoE verifier on a distributed cluster equipped with Huawei Ascend NPUs, using coding-agent trajectories from CodeForge-Preview~\cite{CoderForge2026}. Our preliminary evaluation shows that the proposed post-training substantially improves the model's ability to discriminate between successful and failed trajectories, increasing classification accuracy from 57\% for the base model to 78\% after reinforcement learning on the held-out test set. We open-source \swecritixws on GitHub~\cite{swe_critix} and Hugging Face~\cite{swe_critix-hf}.
\section{SWE-Critix}
Figure~\ref{fig:overview} shows the post-training pipeline of \swecritix, which learns an LLM (i.e., a generalist verifier) for evaluating coding-agent trajectories. The verifier takes an issue-solving trajectory as input and determines whether the issue has been successfully resolved. Before making the binary judgment, the verifier produces a chain-of-thought (CoT) rationale that casually infers its decision (Figure~\ref{fig:overview}-Application). 

To train the verifier, \swecritixws first applies a more capable teacher LLM to review a pair of successful and failed trajectories that address the same issue. Through differential analysis, the teacher summarizes the respective procedures, mistake patterns, and key factors leading to the success or failure of each trajectory (Figure~\ref{fig:overview}-CoT Annotation). These summaries are then incorporated into a small amount of SFT training samples, providing supervision for the base model to learn the reasoning process underlying trajectory evaluation (Figure~\ref{fig:overview}-SFT). Finally, RL is applied to a substantial number of samples without CoT annotations, where unit-testing results are used as ground truth for computing the reward signal (Figure~\ref{fig:overview}-RL).

\subsection{Data Collection}\label{sec:SWE-Critix-Data-Collection}
We collect test-verified trajectories from CoderForge-Preview~\cite{CoderForge2026}, a large open dataset containing 258K trajectories (155K pass vs. 103K fail) spanning 51K issues across 1,655 repositories. These trajectories were generated using the OpenHands scaffold~\cite{wang2025openhandsopenplatformai} with Qwen3-Coder-480B~\cite{qwen3technicalreport} as the underlying model, on three downstream task sources~\cite{jain2025r2egymproceduralenvironmentshybrid,badertdinov2025swerebenchautomatedpipelinetask,yang2025swesmithscalingdatasoftware} that provide executable gym environments. We further prune the dataset by discarding any trajectory that exceeds 64K tokens or does not contain a submitted patch (e.g., the task is incomplete). After filtering out 16\% of the data, we split the remaining trajectories into SFT, RL, and test sets as follows:

\begin{enumerate} [label=\arabic*., leftmargin=*, itemsep=2pt]
	\item We randomly choose 20\% of the issues and use all associated trajectories  as the test set.
	\item Among the remaining issues, we randomly select one successful and one failed trajectory from each issue that contains both types, and assign them to the SFT set.
	\item From the remaining trajectories, we include all failed trajectories in the RL set and then sample successful trajectories in a round-robin fashion across issues until the two classes are balanced. This maximizes the coverage of distinct issues.
\end{enumerate}

 The resulting data is released on Hugging Face~\cite{swe_critix-hf} and its statistics is summarized in Figure~\ref{fig:overview}-Data Splits.

\subsection{Chain-of-Thought (CoT) Annotation}
It has been widely acknowledged that generating a CoT before providing the final answer significantly improves the performance of LLMs to resolve complex tasks~\cite{wei2023chainofthoughtpromptingelicitsreasoning,guan2025deliberativealignmentreasoningenables,chen2026rmr1rewardmodelingreasoning}. To bootstrap the reasoning capabilities of our target verifier, \swecritixws leverages a teacher LLM (GPT-5.5) to generate a structured CoT trace by comparatively analyzing each pair of successful and failed trajectories in the SFT set. We adopt this pairwise comparison approach based on prior findings that LLMs perform better in pairwise evaluations than in individual judgments~\cite{wettig2024quratingselectinghighqualitydata,ouyang2022traininglanguagemodelsfollow}. We then post-process the generated CoTs to adapt them for pointwise judgment, which is the intended evaluation setting for the trained verifier. The prompt templates and more details are released on GitHub~\cite{swe_critix}. The resulting CoT trace of a trajectory includes a concise summary of its issue-resolving procedure, a characterization of mistake patterns  based on a predefined taxonomy, and a causal summary of the factors contributing to the success or failure of the trajectory. We apply CoT annotation only to the SFT set, which contains 10,390 trajectories and accounts for 8.7\% of the training set.

\subsection{Supervised Fine-Tuning (SFT)}
We assemble the annotated CoTs into SFT samples using the templates shown in Figure~\ref{fig:overview}-SFT, where the green and orange boxes represent the templates applied on successful and failed trajectories, respectively. We retain the thinking template of Qwen, as we choose to fine-tune from Qwen3-30B-A3B-Thinking-2507 that already possesses some reasoning capabilities~\cite{qwen3technicalreport}. We minimize the negative log-likelihood of the responses over the SFT set, which is the standard SFT cross-entropy loss:
\begin{equation}
\mathcal{L}_{\mathrm{SFT}}(\theta)
=
-\mathbb{E}_{(x,y)\sim\mathcal{D}_{\mathrm{SFT}}}
\left[
\sum_{t=1}^{|y|}
\log \pi_{\theta}(y_t \mid x, y_{<t})
\right]
\label{eq:sft_loss}
\end{equation}
where $x$ denotes prompts and $y$ denotes the corresponding response, including both the CoT and the final judgment.

\begin{figure}[t]
    \centering
    \includegraphics[width=\columnwidth]{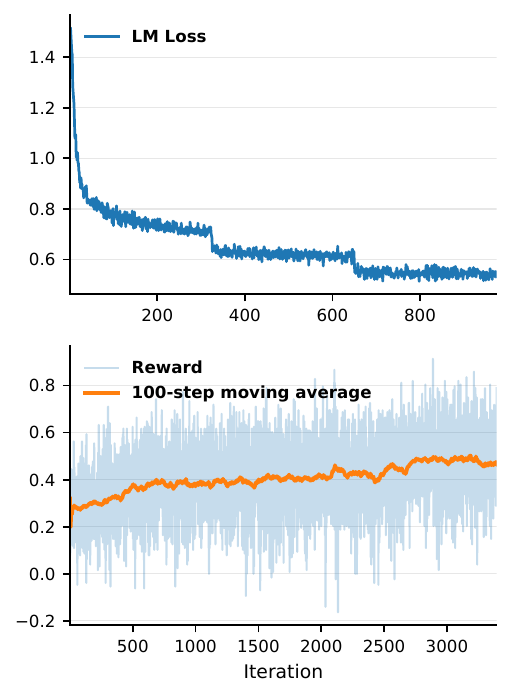}
    \caption{Training dynamics of \swecritix. The upper plot shows the LM loss during SFT, while the lower plot shows the reward during RL training.}
    \label{fig:training_curve}
\end{figure}

We use MindSpeed-LLM~\cite{mindspeed} to fine-tune the Qwen3 MoE model on 128 Huawei Ascend 910B1/B2-64GB NPUs across 16 nodes. Training for three epochs (975 iterations) takes approximately 15 hours. The SFT loss curve is illustrated in Figure~\ref{fig:training_curve}. Training code and hyperparameter settings are available on GitHub~\cite{swe_critix}. We denote the checkpoint at iteration 975 as $\pi_{\mathrm{SFT}}$ and open-source its weights on Hugging Face~\cite{swe_critix-hf}.

\begin{table*}[t]
    \centering
    \caption{Performance comparison of Qwen3-30B-A3B, \swecritixws SFT, and \swecritixws RL on the test set and its three subsets.}
    \label{tab:results}

    \renewcommand{\arraystretch}{1.15}
    \setlength{\tabcolsep}{0pt}

    \begin{tabularx}{\textwidth}{
        @{}
        p{2.0cm}
        p{2.2cm}
        p{2.5cm}
        p{2.5cm}
        p{2.5cm}
        p{0.8cm}
        p{2.4cm}
        p{2.4cm}
        X
        @{}
    }
        \toprule
    
    
        \rowcolor{headerblue}
        \textbf{Dataset}
        & \textbf{Model}
        & \textbf{Accuracy}
        & \textbf{Precision+}
        & \textbf{Recall+}
        & \textbf{F1+}
        & \textbf{Precision-}
        & \textbf{Recall-}
        & \textbf{F1-}
        \\
    
        \midrule
    
    
        \multirow{3}{*}{\textbf{R2E-Gym}}
        & Qwen3-30B-A3B
        & 0.60 (3495/5846) & 0.68 (3483/5128) & 0.88 (3483/3965) & 0.77 & 0.80 (12/15) & 0.01 (12/1881) & 0.01 \\
    
        &
        SWE-Critix SFT
        & 0.52 (3027/5846) & 0.73 (1817/2488) & 0.46 (1817/3965) & 0.56 & 0.36 (1210/3358) & 0.64 (1210/1881) & 0.46 \\

        \rowcolor{rlorange}
        \cellcolor{white}
        &
        SWE-Critix RL
        & 0.62 (3646/5846)
        & 0.75 (2671/3577)
        & 0.67 (2671/3965)
        & 0.71
        & 0.43 (975/2269)
        & 0.52 (975/1881)
        & 0.47
        \\
    
        \midrule
    
    
        \multirow{3}{*}{\textbf{SWE-Rebench}}
        & Qwen3-30B-A3B
        & 0.56 (6675/11872) & 0.65 (6656/10170) & 0.86 (6656/7774) & 0.74 & 0.68 (19/28) & 0.00 (19/4098) & 0.01 \\
    
        &
        SWE-Critix SFT
        & 0.71 (8476/11872) & 0.84 (5369/6360) & 0.69 (5369/7774) & 0.76 & 0.56 (3107/5512) & 0.76 (3107/4098) & 0.65 \\

        \rowcolor{rlorange}
        \cellcolor{white}
        &
        SWE-Critix RL
        & 0.78 (9308/11872)
        & 0.87 (6158/7106)
        & 0.79 (6158/7774)
        & 0.83
        & 0.66 (3150/4766)
        & 0.77 (3150/4098)
        & 0.71
        \\
    
        \midrule
    
    
        \multirow{3}{*}{\textbf{SWE-Smith}}
        & Qwen3-30B-A3B
        & 0.56 (14417/25727) & 0.66 (14397/21793) & 0.85 (14397/17002) & 0.74 & 0.80 (20/25) & 0.00 (20/8725) & 0.00 \\
    
        &
        SWE-Critix SFT
        & 0.66 (17018/25727) & 0.76 (12154/16015) & 0.71 (12154/17002) & 0.74 & 0.50 (4864/9712) & 0.56 (4864/8725) & 0.53 \\

        \rowcolor{rlorange}
        \cellcolor{white}
        &
        SWE-Critix RL
        & 0.82 (20971/25727)
        & 0.82 (15670/19094)
        & 0.92 (15670/17002)
        & 0.87
        & 0.80 (5301/6633)
        & 0.61 (5301/8725)
        & 0.69
        \\
    
        \midrule
    
    
        \multirow{3}{*}{\textbf{All}}
        & Qwen3-30B-A3B
        & 0.57 (24587/43445) & 0.66 (24536/37091) & 0.85 (24536/28741) & 0.75  & 0.75 (51/68) & 0.00 (51/14704) & 0.01 \\
    
        &
        SWE-Critix SFT
        & 0.66 (28521/43445) & 0.78 (19340/24863) & 0.67 (19340/28741) & 0.72 & 0.49 (9181/18582) & 0.62 (9181/14704) & 0.55 \\

        \rowcolor{rlorange}
        \cellcolor{white}
        &
        SWE-Critix RL
        & 0.78 (33925/43445)
        & 0.82 (24499/29777)
        & 0.85 (24499/28741)
        & 0.84
        & 0.69 (9426/13668)
        & 0.64 (9426/14704)
        & 0.66
        \\
        \bottomrule
    \end{tabularx}
\end{table*}

\subsection{Reinforcement Learning (RL)}
The SFT stage utilizes a small fraction of the training data with CoT annotations to help the model learn the reasoning patterns before making the final judgment.To further enhance the model's reasoning ability, we apply RL, a more powerful learning paradigm~\cite{deepseekai2025deepseekr1incentivizingreasoningcapability,shao2024deepseekmathpushinglimitsmathematical,lambert2025reinforcementlearninghumanfeedback}, to the remaining 91.3\% of the training data, which contain unit-testing results but no CoT annotations. We follow the standard policy gradient algorithms to maximize the expected rewards obtained by a policy $\pi_{\theta}$:
\begin{equation}
\max_{\pi_{\theta}}\;
\mathbb{E}_{x\sim D_{\mathrm{RL}}, y\sim\pi_\theta(.|x)}
\left[r(x,y)\right]
-
\beta \mathbb{D}_{\mathrm{KL}}
\left[
\pi_\theta \,\|\, \pi_{\mathrm{SFT}}
\right]
\label{eq:rl_obj}
\end{equation}
where the regularization term $\mathbb{D}_{\mathrm{KL}}$ prevents $\pi_\theta$ from deviating too far from $\pi_{\mathrm{SFT}}$.

We define a simple yet effective rule-based outcome reward:
\begin{equation}
r(x,y) =
\begin{cases}
+1, & \text{if } y_{\mathrm{ans}} = \text{YES} \land x_{\mathrm{traj}} \text{ passes tests}\\
+1, & \text{if } y_{\mathrm{ans}} = \text{NO} \land x_{\mathrm{traj}} \text{ fails tests}\\
-1, & \text{otherwise}.
\end{cases}
\label{eq:rl_reward}
\end{equation}
where $x_{\mathrm{traj}}$ denotes the coding-agent trajectory in prompt $x$ and $y_{\mathrm{ans}}$ denotes the final answer extracted from the response $y$. A response receives a reward of +1 if its predicted judgment is consistent with the ground truth derived from unit testing, and -1 otherwise.

We use $\pi_{\mathrm{SFT}}$ as the initial RL checkpoint and conduct RL training using 128 Huawei Ascend 910C-64GB NPUs across 8 nodes with VeRL~\cite{Sheng_2025}. We use GRPO~\cite{shao2024deepseekmathpushinglimitsmathematical} with group size as 8 to optimize the objective in Equation~\ref{eq:rl_obj}. One epoch consists of 3,396 iterations and takes approximately 533 hours to complete. The reward curve during training is shown in Figure~\ref{fig:training_curve}. We refer to the resulting checkpoint as $\pi_{\mathrm{RL}}$ and release its weights on Hugging Face~\cite{swe_critix-hf}.

\section{Preliminary Evaluation}

As a preliminary evaluation, we focus on the model's ability to distinguish between successful and failed trajectories. We evaluate $\pi_{\mathrm{RL}}$, $\pi_{\mathrm{SFT}}$, and the base Qwen3 MoE model on the test split created in Section~\ref{sec:SWE-Critix-Data-Collection}. We serve these models using vLLM Ascend~\cite{vllm_ascend} on one node containing 16 Huawei Ascend 910C-64GB NPUs. Given the large scale of the test set (over 43K long-trajectory samples), we evaluate each model on each sample only once. We report \textit{accuracy}, as well as \textit{precision}, \textit{recall}, and \textit{F1} scores for both the positive (successful trajectory) and negative (failed trajectory) classes.

We present the results in Table~\ref{tab:results}. The results reveal a clear progression from general-purpose language modeling to specialized trajectory verification. The base Qwen3 MoE model exhibits a strong bias toward predicting trajectories as correct, as its negative recall is nearly zero across all subsets. It also has limited instruction following capability on this task. We observed that 9.6\% of its responses do not adhere to the required output format, preventing reliable extraction of the predicted judgments. SFT substantially alleviates both issues. All responses produced by $\pi_{\mathrm{SFT}}$ strictly follow the instructions required by the prompt, and the model becomes considerably less biased toward positive predictions. Further RL training yields another significant improvement, increasing accuracy from 66\% after SFT to 78\% on the overall test set. These results suggest that SFT primarily addresses the base model's strong positive bias, while subsequent RL further strengthens its discriminative capability by recovering positive recall from 67\% to 85\% without substantially sacrificing negative recall.

Despite these gains, two limitations remain. First, performance on R2E-Gym is notably lower than those on SWE-Rebench and SWE-Smith. One possible explanation is its relatively smaller data volume. Second, $\pi_{\mathrm{RL}}$ remains less effective at recognizing failed trajectories as the negative recall reaches only 64\% overall. This suggests that failed trajectories remain more challenging to distinguish from successful ones, potentially due to the greater diversity of failure patterns and the subtle failure signals that are difficult to identify.


\section{Future Work}
Our current evaluation focuses on the discriminative ability of \swecritixws. In future, we plan to conduct more comprehensive evaluations along the following five dimensions.

\textbf{1. Training Data}: All samples in CodeForge-Preview~\cite{CoderForge2026} are generated using OpenHands~\cite{wang2025openhandsopenplatformai} with Qwen3-Coder-480B~\cite{qwen3technicalreport}. To investigate the impact of data mixtures, we plan to incorporate trajectories from other sources, such as Open-SWE-Traces~\cite{ahmad2026openswetracesadvancingdualmodemultilingual}, and synthesize trajectories using different combinations of LLMs and harnesses. Furthermore, beyond the balanced positive-to-negative ratio, we plan to investigate the effects of different class ratios, such as 2:1, 1:2, and 4:1.

\textbf{2. Training Methods}: It remains an open question whether RL alone is sufficient to train an effective verifier. We therefore plan to establish a cold-start RL baseline by removing the CoT annotation and SFT stages from our pipeline. In addition, we plan to conduct reward ablations in the RL stage by incorporating signals not only from the final test results but also from the intermediate reasoning process.

\textbf{3. Downstream Applications}: We plan to explore downstream scenarios that can benefit from the trained verifier. For example, can the verifier help recover a correct solution by identifying and rectifying errors in a failed trajectory? Can the verifier serve as a proxy reward for RL training when ground-truth test results are unavailable?

\textbf{4. Baselines}: We plan to compare \swecritixws with frontier open/closed-source models~\cite{deepseekai2026deepseekv41flash,glm5team2026glm5vibecodingagentic,gpt-6-astra} and representative methods from the categories summarized in Table~\ref{tab:verifier_taxonomy}.

\textbf{5. Qualitative Analysis}: Beyond the quantitative evaluations described above, we plan to conduct a qualitative analysis of the model's reasoning traces to assess whether its analyses are logically coherent and grounded in evidence. We will also investigate cases in which the verifier makes incorrect judgments to better understand the sources of its errors.
\section{Related Work}
Prior work proposing verifiers for reviewing coding-agent executions can be classified into two major categories depending on whether a specialized verification model is trained. One is training-free verification, where an evaluator agent uses general-purpose LLMs to assess coding agent executions through test generation~\cite{xia2024agentlessdemystifyingllmbasedsoftware,traeresearchteam2025traeagentllmbasedagent}, rubric-based grading~\cite{raghavendra2026agenticrubricscontextualverifiers,gandhi2025agentsastraycoursecorrectingswe,nanda2026winkrecoveringmisbehaviorscoding}, verification scaling~\cite{kwok2026llmasaverifiergeneralpurposeverificationframework}, metamorphic  relations~\cite{li2026independentpatchverificationcoding}, failure attribution~\cite{qi2026trajdebugtracingerrorlifecycle}, or anti-pattern detection~\cite{zheng2026sweprimefewertrajectoriesbetter,liu2026onlinemonitoringcorrectivesteering,sahoo2026agentlensrevealingluckypass,he2026procctrlbenchevaluatingprocessleveldefects}. Another major category is training-based verification, where a specialized critic is trained to evaluate trajectories using different learning paradigms such as TD learning~\cite{wang2025-sota-on-swe-bench,zainullina2025guidedsearchstrategiesnonserializable}, autoregressive next-token prediction~\cite{pan2025trainingsoftwareengineeringagents,jain2025r2egymproceduralenvironmentshybrid,deepswe2025,shum2025swermexecutionfreefeedbacksoftware,zeng2026dockerlessenvironmentfreeprogramverifier,tao2026execcriticlearntesttest,raghavendra2026agenticrubricscontextualverifiers,gandhi2026steerdontsolvetraining,wang2026swereviewclosingloopissue,rahman2026locatinghiddenfailuresmakes}, multi-label classification~\cite{wang2026rubricsupervisedcriticsparserealworld}, DPO~\cite{gandhi2026steerdontsolvetraining}, Bradley-Terry~\cite{han2026swetraceoptimizinglonghorizonswe}, and policy gradient~\cite{tao2026execcriticlearntesttest,rahman2026locatinghiddenfailuresmakes}.
\section{Conclusion}
In this paper, we proposed a post-training pipeline \swecritixws that integrates SFT with RL to learn a process-level generalist verifier for assessing coding-agent trajectories. Preliminary experiments on coding-agent trajectories from CodeForge-Preview demonstrate a substantial improvement in the model's discriminative ability.


\bibliographystyle{IEEEtran}
\balance
\bibliography{references}

\end{document}